%% file: main.tex
\documentclass[10pt,twocolumn,letterpaper]{article}

\usepackage[pagenumbers]{iccv} 


\definecolor{iccvblue}{rgb}{0.21,0.49,0.74}
\usepackage[pagebackref,breaklinks,colorlinks,allcolors=iccvblue]{hyperref}
\usepackage{multirow}
\usepackage{makecell}

\def\paperID{12459} 
\def\confName{ICCV}
\def\confYear{2025}

\title{Self Distillation via Iterative Constructive Perturbations}

\author{
\begin{tabular}{ccc}
Maheak Dave$^{1}$ & Aniket Kumar Singh$^{1}$ & Aryan Pareek$^{1}$ \\
Harshita Jha$^{1}$ & Debasis Chaudhuri$^{1}$ & Manish Pratap Singh$^{2}$ \\
\multicolumn{3}{c}{
\small $^1$Techno India University \quad \small $^2$DRDO Young Scientist Laboratory - Cognitive Technologies
}
\end{tabular}
}

\begin{document}
\maketitle
\input{sec/0_abstract}    
\input{sec/1_intro}
\input{sec/2_background}
\input{sec/3_methodology}
\input{sec/4_experiments}
\input{sec/5_conclusion}
{
    \small
    \bibliographystyle{ieeenat_fullname}
    \bibliography{main}
}

\end{document}

%% file: sec/0_abstract.tex
\begin{abstract}
Deep Neural Networks have achieved remarkable achievements across various domains, however balancing performance and generalization still remains a challenge while training these networks. In this paper, we propose a novel framework that uses a cyclic optimization strategy to concurrently optimize the model and its input data for better training, rethinking the traditional training paradigm. Central to our approach is Iterative Constructive Perturbation (ICP), which leverages the model’s loss to iteratively perturb the input, progressively constructing an enhanced representation over some refinement steps. This ICP input is then fed back into the model to produce improved intermediate features, which serve as a target in a self-distillation framework against the original features. By alternately altering the model's parameters to the data and the data to the model, our method effectively addresses the gap between fitting and generalization, leading to enhanced performance. Extensive experiments demonstrate that our approach not only mitigates common performance bottlenecks in neural networks but also demonstrates significant improvements across training variations.
\end{abstract}

%% file: sec/1_intro.tex
\section{Introduction}
\label{sec:intro}
Over the past decade, deep learning models have yielded outstanding performance in many areas, including computer vision, natural language processing, healthcare, and autonomous systems \cite{lecun2015deep,krizhevsky2012imagenet,devlin2019bert}. Despite such successes, optimizing model performance, especially for real-world cases with heterogeneous input distributions, is an ongoing research topic \cite{zhang2021understanding}. The systematic adjustment or modification of input data with the aim of improving prediction accuracy (sometimes referred to as input optimization) has emerged as a key strategy for enhancing model performance \cite{smith2017bayesian}.

Classic performance improvement techniques have centered on architectural advancements, training methods, and hyperparameter optimization \cite{he2016deep,ioffe2015batch}, with encouraging outcomes in accuracy and generalization. Recent developments in ensemble techniques \cite{lakshminarayanan2017simple}, feature space manipulation \cite{wang2018esrgan,furlanello2018born}, and auxiliary learning tasks \cite{DBLP:journals/corr/abs-1708-07860,gidaris2018unsupervised} emphasize the importance of rich, informative representations. Yet, most of these methods are accompanied by higher computational complexity, with room for more efficient optimization methods.

Self-distillation methods have become potent means of enhancing model performance. Following seminal research in knowledge distillation \cite{hinton2015distilling}, conventional self-distillation methods as in \cite{furlanello2018born,DBLP:journals/corr/abs-1812-00123}, often consist of training subsequent generations of models where every generation is trained on the previous one. Approaches such as BYOT \cite{8578552} involve additional parameters for producing soft labels for self-distillation, imposing computational overhead and complexity on learning. Likewise, methods like CS-KD \cite{DBLP:journals/corr/abs-2003-13964} are mainly concerned with output logits, and they may lose important information present in intermediate feature representations.

State-of-the-art optimization methods in deep learning have been more geared towards weight optimization than enhancing input representation. Recent work \cite{NEURIPS2020_2288f691} illustrates that enhancing feature quality at training time is at the core of model performance and generalization. The problem is to create techniques that can improve input representations systematically while optimizing model performance \cite{DBLP:journals/corr/abs-2104-02057}.

To tackle these issues, we introduce a new framework integrating Iterative Constructive Perturbation (ICP) and self-distillation. ICP optimizes input representations by iterative gradient-based updates, using the step size as a dynamic learning rate for accurate input optimization. Supplementing ICP, our self-distillation approach synchronizes feature representations between original and optimized inputs in a unified training process. By using a cosine decay weighting scheme, we prioritize various network layers during training, successfully capturing both basic and abstract features.

This method moves the emphasis from standard weight optimization to anticipatory input refinement, providing a computationally effective means to improve performance. Through the combination of ICP and self-distillation, our research provides a systematic approach to enhancing feature quality and neural network accuracy, building on current attempts to optimize deep learning models with novel learning methods.

In conclusion, this paper offers a new method that integrates Iterative Constructive Perturbations (ICP) into a self-distillation process to improve model performance and accuracy. Through the integration of ICP with a feature alignment strategy, our approach applies constructive input perturbations systematically to enhance feature quality and overall performance. This method changes the focus to proactive model improvement through input optimization. The coupling of ICP and self-distillation provides a systematic approach toward gaining better model representations, helping to further the debate on maximizing neural network performance through new perturbation and learning methods.

%% file: sec/2_background.tex
\section{Theoretical Background}
\subsection{FGSM (Fast Gradient Sign Method)}
Fast Gradient Sign Method (FGSM) \cite{goodfellow2015explainingharnessingadversarialexamples} is a gradient-based adversarial attack that adds small, targeted perturbations to the input data ($x$, represented as a vector or tensor of feature values) in the direction of the gradient of the loss function $J(\theta,x,y)$ with respect to $x$. Formally, the perturbed input is given by:
 \begin{equation}
    x^{adv} = x + \epsilon \cdot sgn(\nabla_x J(\theta,x,y))
\end{equation}
where $\epsilon$ controls the magnitude of the perturbation, $\nabla_x J$ represents the gradient of the loss function with respect to the input, and $sgn(\cdot)$ ensures the perturbation is applied uniformly in the gradient's direction while also constraining its magnitude. This constraint prevents the perturbation from becoming excessively large, making it imperceptible to human observers while still being effective in misleading the model. By leveraging directional information through gradients, FGSM efficiently perturbs the input to maximize the model's loss with minimal visible alteration.
\subsection{Iterative Constructive Perturbation (ICP)}
This paper introduces a novel approach to input refinement using ICP, drawing inspiration from the gradient-based logic of FGSM. While FGSM generates adversarial examples by applying a single-step perturbation to maximize the model's loss, ICP reverses this concept, iteratively refining inputs to minimize the loss and enhance performance. By employing multiple iterations of gradient-based adjustments, ICP systematically aligns inputs with the model’s learned features, offering a more robust and effective strategy for input optimization. The method can be expressed as:
\begin{equation}
    x_t=x_{t-1}-\epsilon \cdot\nabla_{x_{t-1}} J(\theta,x_{t-1},y)
\end{equation}
where:
    \begin{itemize}
        \item $x_t$ is the output at iteration $t$
        \item $x_{t-1}$ is the output from previous iteration $t-1$
        \item $x_0$ is the original input
    \end{itemize}
    
FGSM uses only the sign of the gradient, resulting in uniform perturbations for pixels with the same gradient sign. In contrast, ICP leverages the full gradient information, producing perturbations proportional to the exact gradient value at each pixel. This approach allows for more nuanced adjustments, with iterative steps further refining inputs to align with the model's learned features.
\begin{figure} [t]
    \centering
    \includegraphics[width=\linewidth]{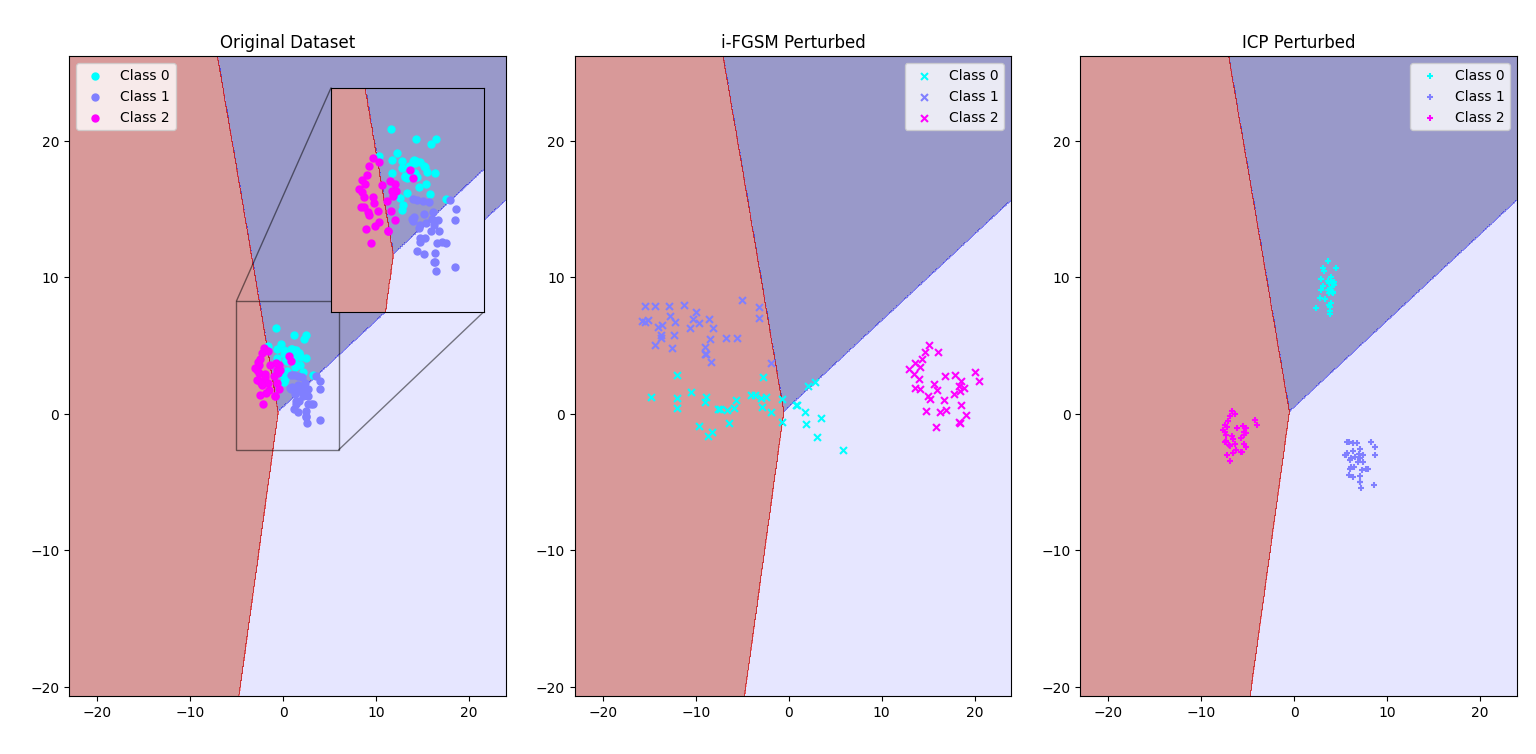}
    \caption{Plot depicting effects of ICP and i-FGSM in a simple multi-class classification scenario}
    \label{fig:ICP vs iFGSM}
\end{figure}

Figure \ref{fig:ICP vs iFGSM} illustrates the effect of ICP and iterative FGSM (i-FGSM), which is simply an iterative version of FGSM, with $\epsilon = 0.002$ for both the perturbation techniques, on a dataset of three isotropic Gaussian clusters, classified using a simple multi-layer perceptron (MLP). Each cluster, shown with distinct coloured markers (Original Class 0–2), represents a separate class.

The leftmost panel displays the original dataset along with the decision boundary learned by the MLP. The middle panel presents the effect of i-FGSM perturbation, where samples are iteratively modified to maximize the model's loss. This leads to a significant shift in data points across decision boundaries, demonstrating the adversarial nature of the attack. As a result, many perturbed samples are misclassified, highlighting the model's vulnerability to adversarial perturbations.

In comparison, the rightmost panel shows the effect of ICP perturbation, where samples are iteratively refined to reduce model loss while preserving feature consistency. Unlike i-FGSM, which tends to push them out of their original associated regions, increasing the chances of misclassification, ICP preserves each cluster’s alignment by shifting points further from the boundary while keeping them within their natural regions. As a result, ICP enhances class separation without disrupting the underlying data structure, leading to improved overall robustness.

\subsection{Self Distillation}
Self-distillation is a streamlined variation of knowledge distillation where a model learns from its own predictions rather than relying on a larger teacher model. Unlike traditional knowledge distillation, which trains a smaller "student" model to replicate the outputs of a pre-trained "teacher," self-distillation either trains a new iteration of the same model using its previous predictions or refines predictions during training.

This process leverages "soft" labels probabilistic outputs from earlier iterations or intermediate layers allowing the model to capture nuanced data patterns, improve generalization, and reduce overfitting. By learning from soft labels, self-distillation smooths decision boundaries, mitigating overconfidence and addressing class imbalance, even in noisy or adversarial settings. Furthermore, it accelerates convergence and simplifies the training process by eliminating the need for a separate teacher model, making it an efficient choice for resource-constrained environments.

%% file: sec/3_methodology.tex
\begin{figure} [t]
    \centering
    \includegraphics[width=\linewidth]{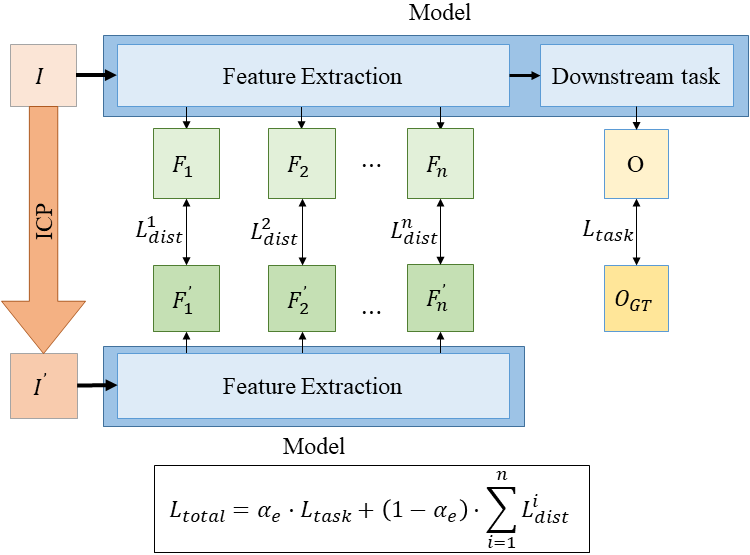}
    \caption{Overview of the proposed ICP based self-distillation framework}
    \label{fig:framework}
\end{figure}
\input{tables/hyperparam_search}
\section{Proposed Methodology}
\label{sec:methodology}

Our approach utilizes ICP within a self-distillation framework to enhance model performance through feature alignment across intermediate layers. This methodology includes an initial baseline training phase for a total of $k$ epochs, followed by self-distillation with dynamic adjustments in layer-wise and loss weights to refine feature representations effectively. The proposed training scheme begins with a neural network designed for a specific task, typically consisting of two components: a feature extraction module followed by a downstream task (e.g., classification). Figure \ref{fig:framework} gives a brief overview of the proposed methodology.
\subsection{Baseline Training Phase}
For initial $k$ baseline epochs, the model is trained using regular training scheme for its specific downstream task and is optimised using the task specific loss $L_{task}$ , establishing foundational representations to ensure accurate task performance.
\subsection{ICP-Driven Self-Distillation Phase}
\subsubsection{ICP Adjustment}
A forward pass is performed using the original input $I$, yielding the task-specific output $O$, intermediate feature maps $F_i$, and the corresponding task-specific loss $L_{task}$. Here, $i=1$ to $n$ with $n$ being the number of layers of the model selected for distillation. $L_{task}$ is then utilized by the ICP process to generate a perturbed input $I'$ through iterative optimization over $T$ iterations. Adjusting the input in the gradient's negative direction to boost task performance. This adjustment improves the model's performance by guiding the input more closely within the feature space learned by the model.
\subsubsection{Alignment of Intermediate Feature Representations}
The modified input $I'$ is passed through the network again to obtain updated intermediate feature maps $F_i'$. Layer-wise distillation losses $L_{dist}^i$ are computed by comparing feature maps $F_i$ with $F_i'$. In this paper, the mean-squared error (MSE) loss has been utilised as the distillation loss metric such that:
\begin{equation}
    L_{dist}^i = \text{MSE}\left( F_i, F_i' \right)
\end{equation}
We also consider a weighted combination of the individual layer wise loss, where weighing function should be a strictly increasing function to give more weightage to deeper layers, 
\subsection{Combined Training Loss with Cosine-Decay}
For the initial $k$ baseline epochs, there is no distillation. Only after $k$ baseline epochs, the self-distillation phase begins. The balance between task specific loss $L_{task}$ and distillation losses $L_{dist}^i$ is controlled by a parameter $\alpha_e$, which evolves with each epoch $e$.  The total loss is defined as:
\begin{equation}
    L_{total}=\alpha_e \cdot L_{task} + (1-\alpha_e)\cdot \sum_{i=1}^nL_{dist}^i
\end{equation}
with $\alpha_e$ calculated as:
\begin{equation}
    \alpha_e = \begin{cases}
    1 & \text{, } e \le k\\
      \cos\left(\frac{\pi(e-k)}{2(E-k)}\right) & \text{, } e>k
    \end{cases}
\end{equation}
Here, $E$ is the total number of epochs for training. For baseline epochs, $\alpha_e=1$ results in no weightage for the self-distillation task. After $k$ epochs, $\alpha_e$ follows cosine-decay scheduling method, inspired by principles in learning rate decay strategies. Thus, the use of the parameter $\alpha_e$ optimally balances between task performance and robust feature representation by adjusting the model's focus over time.
In summary, the framework processes both the original input $I$ and the ICP-modified input $I'$, aligning feature representations through layer-wise distillation losses. By leveraging ICP without a fixed perturbation limit, this approach effectively aligns input with learned features. The cosine-decayed self-distillation approach further smoothens model convergence, enhancing performance and efficiency for real-world applications.
\subsection{ICP Variants}
Since Iterative Constructive Perturbation (ICP) is basically a process of gradient descent on input samples, it is possible to extend it based on contemporary optimization techniques. The base ICP is here on termed as SGD-ICP. To see the influence of other modern optimization algorithms based ICP alternatives, we considered two alternatives: Adam-ICP and AdEMAMix-ICP. These alternatives take their name after the Adam \cite{kingma2014adam} and AdEMAMix \cite{pagliardini2024ademamix} optimization techniques, respectively, and are motivated by the potential in improving the effectiveness of ICP.
\subsubsection{Adam-ICP}
Adam-ICP incorporates the Adam optimization algorithm into the ICP algorithm by keeping first-order and second-order moment estimates of gradients. The first moment vector $m_t$ is the exponentially moving average of previous gradients, and the second moment vector $v_t$ regards the squared gradients for adaptive scaling. The update equations for Adam-ICP are:
\begin{equation}
    m^{(t)} =\beta_1 m^{(t-1)} +(1-\beta_1) \nabla_{x_{t-1}}L
\end{equation}
\begin{equation}
    v^{(t)} =\beta_2 v^{(t-1)} +(1-\beta_2) (\nabla_{x_{t-1}}L)^2
\end{equation}
\begin{equation}
    x_t = x_{t-1} - \epsilon \cdot \frac{m^{(t)} \sqrt{1-\beta_2^t}}{(1-\beta_1^t)\sqrt{v^{(t)}}}
\end{equation}
This formulation applies the Adam-like approach of scaling updates by first- and second-moment gradient estimates to achieve more efficient perturbation steps and more stable convergence in the ICP context.
\subsubsection{AdEMAMix-ICP}
AdEMAMix-ICP adds the AdEMAMix strategy to the Adam-ICP formulation, which adds an extra momentum term $m_2^t$ for improved variance control. This variance improves stability through blending various amounts of moment estimates, rendering it stronger against sudden gradient fluctuations. The update equations are:
\begin{equation}
    m_1^{(t)} =\beta_1 m_1^{(t-1)} +(1-\beta_1) \nabla_{x_{t-1}}L
\end{equation}
\begin{equation}
    m_2^{(t)} =\beta_3^{(t)} m_2^{(t-1)} +(1-\beta_3^{(t)}) \nabla_{x_{t-1}}L
\end{equation}
\begin{equation}
    x_t = x_{t-1} - \epsilon \cdot \frac{(m_1^{(t)} + \alpha^{(t)} m_2^{(t)} (1-\beta_1^t)) \sqrt{1-\beta_2^t}}{(1-\beta_1^t)\sqrt{v^{(t)}}}
\end{equation}
These ICP variants facilitate a more optimizer-driven and flexible perturbation generation, offering enhanced generalization and stability in various learning situations.

%% file: tables/hyperparam_search.tex
\begin{table*}[t]
    \centering
    \small
    \begin{subtable}{0.45\linewidth}
        \begin{tabular}{@{}lcccccc@{}}
            \toprule
            \textbf{Method} & \textbf{k} & \textbf{T} & \textbf{Weighted} & \makecell{\textbf{Acc.}\\\textbf{(\%)}} & \textbf{F1} & \makecell{\textbf{Time}\\\textbf{(mins)}} \\
            \midrule
            \multirow{16}{*}{\textbf{SGD-ICP}}
                & \multirow{4}{*}{0} 
                    & \multirow{2}{*}{5} 
                    & False & 40.23 & 0.399 & 39.84 \\
                & 
                    & 
                    & True  & 39.91 & 0.393 & 39.88 \\
                \cmidrule(lr){3-7}
                &
                    & \multirow{2}{*}{10} 
                    & False & 39.68 & 0.387 & 45.98 \\
                & 
                    & 
                    & True  & 39.08 & 0.385 & 46.25 \\
                \cmidrule(lr){2-7}
                & \multirow{4}{*}{25} 
                    & \multirow{2}{*}{5} 
                    & False & 40.90 & 0.402 & 36.97 \\
                & 
                    & 
                    & True  & 39.64 & 0.385 & 36.96 \\
                \cmidrule(lr){3-7}
                & 
                    & \multirow{2}{*}{10} 
                    & False & 40.29 & 0.396 & 42.03 \\
                & 
                    & 
                    & True  & 40.60 & 0.400 & 42.03 \\
                \cmidrule(lr){2-7}
                & \multirow{4}{*}{50} 
                    & \multirow{2}{*}{5} 
                    & False & 38.95 & 0.383 & 34.28 \\
                & 
                    & 
                    & True  & 38.83 & 0.377 & 34.38 \\
                \cmidrule(lr){3-7}
                & 
                    & \multirow{2}{*}{10} 
                    & False & 39.11 & 0.389 & 37.64 \\
                & 
                    & 
                    & True  & 38.71 & 0.385 & 37.64 \\
                \cmidrule(lr){2-7}
                & \multirow{4}{*}{75} 
                    & \multirow{2}{*}{5} 
                    & False & 32.25 & 0.313 & 31.46 \\
                & 
                    & 
                    & True  & 31.19 & 0.304 & 31.69 \\
                \cmidrule(lr){3-7}
                & 
                    & \multirow{2}{*}{10} 
                    & False & 34.55 & 0.335 & 33.27 \\
                & 
                    & 
                    & True  & 34.40 & 0.336 & 33.09 \\
            \bottomrule
        \end{tabular}
    \end{subtable}
    \begin{subtable}{0.45\linewidth}
        \begin{tabular}{lcccccc@{}}
            \toprule
            \textbf{Method} & \textbf{k} & \textbf{T} & \textbf{Weighted} & \makecell{\textbf{Acc.}\\\textbf{(\%)}} & \textbf{F1} & \makecell{\textbf{Time}\\\textbf{(mins)}} \\
            \midrule
            \multirow{16}{*}{\textbf{Adam-ICP}}
                & \multirow{4}{*}{0} 
                    & \multirow{2}{*}{5} 
                    & False & 38.59 & 0.378 & 39.55 \\
                & 
                    & 
                    & True  & 37.77 & 0.373 & 39.51 \\
                \cmidrule(lr){3-7}
                &
                    & \multirow{2}{*}{10} 
                    & False & 38.77 & 0.380 & 46.31 \\
                & 
                    & 
                    & True  & 37.70 & 0.370 & 46.06 \\
                \cmidrule(lr){2-7}
                & \multirow{4}{*}{25} 
                    & \multirow{2}{*}{5} 
                    & False & 40.80 & 0.404 & 36.54 \\
                & 
                    & 
                    & True  & 41.31 & 0.409 & 37.48 \\
                \cmidrule(lr){3-7}
                & 
                    & \multirow{2}{*}{10} 
                    & False & 40.39 & 0.399 & 41.98 \\
                & 
                    & 
                    & True  & 41.32 & 0.405 & 41.50 \\
                \cmidrule(lr){2-7}
                & \multirow{4}{*}{50} 
                    & \multirow{2}{*}{5} 
                    & False & 38.39 & 0.374 & 33.40 \\
                & 
                    & 
                    & True  & 36.96 & 0.356 & 34.23 \\
                \cmidrule(lr){3-7}
                & 
                    & \multirow{2}{*}{10} 
                    & False & 39.32 & 0.386 & 37.26 \\
                & 
                    & 
                    & True  & 39.38 & 0.390 & 36.98 \\
                \cmidrule(lr){2-7}
                & \multirow{4}{*}{75} 
                    & \multirow{2}{*}{5} 
                    & False & 29.33 & 0.282 & 31.39 \\
                & 
                    & 
                    & True  & 29.43 & 0.285 & 31.00 \\
                \cmidrule(lr){3-7}
                & 
                    & \multirow{2}{*}{10} 
                    & False & 28.01 & 0.258 & 32.82 \\
                & 
                    & 
                    & True  & 28.88 & 0.272 & 32.97 \\
            \bottomrule
        \end{tabular}
    \end{subtable}
    \begin{subtable}{\linewidth}
        \centering
        \begin{tabular}{lcccccc}
            \textbf{Method} & \textbf{k} & \textbf{T} & \textbf{Weighted} & \makecell{\textbf{Acc.}\\\textbf{(\%)}} & \textbf{F1} & \makecell{\textbf{Time}\\\textbf{(mins)}} \\
            \midrule
            \multirow{16}{*}{\textbf{AdEMAMix-ICP}}
                & \multirow{4}{*}{0} 
                    & \multirow{2}{*}{5} 
                    & False & 38.93 & 0.385 & 40.19 \\
                & 
                    & 
                    & True  & 40.51 & 0.394 & 39.87 \\
                \cmidrule(lr){3-7}
                &
                    & \multirow{2}{*}{10} 
                    & False & 37.82 & 0.370 & 47.90 \\
                & 
                    & 
                    & True  & 36.30 & 0.352 & 46.98 \\
                \cmidrule(lr){2-7}
                & \multirow{4}{*}{25} 
                    & \multirow{2}{*}{5} 
                    & False & 40.80 & 0.399 & 37.36 \\
                & 
                    & 
                    & True  & \textbf{41.99} & \textbf{0.414} & 37.37 \\
                \cmidrule(lr){3-7}
                & 
                    & \multirow{2}{*}{10} 
                    & False & 41.27 & 0.407 & 42.08 \\
                & 
                    & 
                    & True  & 41.43 & 0.405 & 42.32 \\
                \cmidrule(lr){2-7}
                & \multirow{4}{*}{50} 
                    & \multirow{2}{*}{5} 
                    & False & 37.20 & 0.363 & 34.03 \\
                & 
                    & 
                    & True  & 37.80 & 0.373 & 33.81 \\
                \cmidrule(lr){3-7}
                & 
                    & \multirow{2}{*}{10} 
                    & False & 39.53 & 0.391 & 38.02 \\
                & 
                    & 
                    & True  & 26.27 & 0.249 & 37.84 \\
                \cmidrule(lr){2-7}
                & \multirow{4}{*}{75} 
                    & \multirow{2}{*}{5} 
                    & False & 26.27 & 0.249 & 32.44 \\
                & 
                    & 
                    & True  & 26.15 & 0.246 & 31.66 \\
                \cmidrule(lr){3-7}
                & 
                    & \multirow{2}{*}{10} 
                    & False & 27.86 & 0.260 & 33.46 \\
                & 
                    & 
                    & True  & 28.26 & 0.268 & 34.05 \\
            \midrule
            \textbf{Control} & 100 & \textbf{--} & \textbf{--} & 22.93 & 0.229 & 29.25 \\
            \bottomrule
        \end{tabular}
    \end{subtable}
  \caption{Ablation study on SGD-ICP,Adam-ICP and AdEMAMix-ICP with different \(k\), \(T\), and weighting schemes.}
  \label{tab:hyperparam}
\end{table*}

%% file: sec/4_experiments.tex
\section{Experiments}
\subsection{Experimental Setup}
To rigorously evaluate the effectiveness of our proposed ICP based self-distillation framework, we conducted extensive experiments specifically on image classification, and image generation. This multi-task evaluation was designed to test our theory in diverse settings and demonstrate the broad applicability of our method. To ensure a fair and consistent comparison across experiments, all models were trained for a fixed duration of 100 epochs using identical hyperparameters.
\subsection{Hyperparameter Tuning using Image Classification}
We performed experiments for image classification task on the CIFAR-100 \cite{krizhevsky2009learning} dataset using the modified ResNet20 \cite{he2016deep}. These experiments, as observed in Table \ref{tab:hyperparam}, revealed that the optimal configuration for ICP-based self-distillation was achieved at $k = 25$ and $T = 5$ with weighted feature maps for self-distillation. This setting resulted in an improvement in accuracy for all ICP-variants over the control baseline ($k = 100$). Specifically, for \textbf{AdEMAMix-ICP}, the highest recorded accuracy was \textbf{19.06\%} more than the control baseline. In addition, enhancements in F1-score further validated the efficacy of our approach in refining feature representations. Consequently, we adopted these hyperparameters ($k = 25, T = 5, \text{Weighted}=\text{True}$) for all subsequent experiments to ensure consistency across tasks, while also comparing the different ICP variants (SGD-ICP, AdamICP, and AdEMAMix-ICP) to further optimize performance.
\subsection{Image Generation}
To extend our investigation beyond classification, we applied the ICP-based self-distillation framework to image generation task. In this experiment, we evaluated the impact of our framework using a Variational Auto Encoder (VAE) \cite{kingma2013auto} trained on the CUB dataset \cite{wah2011caltech}. The VAE was optimized with the usual task loss used for VAEs: a mix of MSE and KL loss. The model was trained for 100 epochs using the previously found optimal hyperparameters ($k = 25, T = 5, \text{Weighted}=\text{True}$). Evaluation was performed using the Structural Similarity Index (SSIM) and the Fréchet Inception Distance (FID) \cite{heusel2017gans}. We specifically utilized a small dataset and small model, and with images of size $128 \times 128$ to test how our methods work in constrained environments on challenging tasks.
\input{tables/results}
\begin{figure} [ht]
    \centering
    \includegraphics[width=\linewidth]{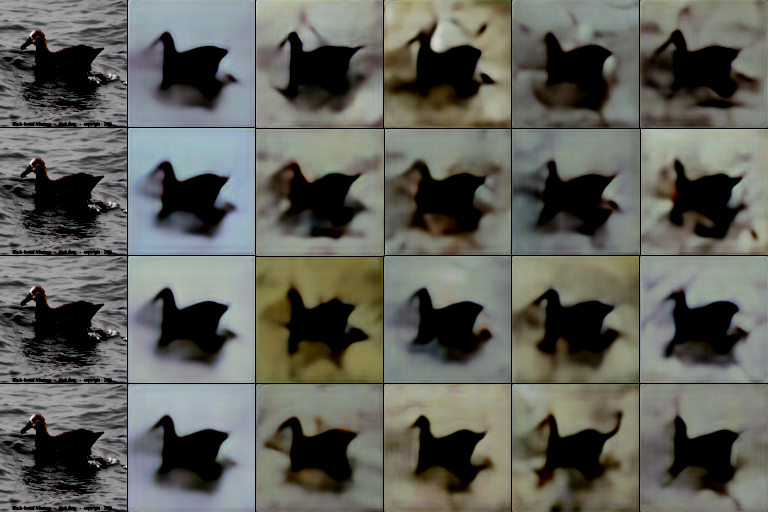}
    \caption{\textbf{Left to right:} Input image from CUB dataset, deterministic output of VAE (with no variance), outputs of VAE with 4 different noised latents with different seeds; \textbf{Top to bottom:} Baseline control method, SGD-ICP, Adam-ICP, and AdEMAMix-ICP.}
    \label{fig:generate}
\end{figure}
\subsection{Results and Analysis}
Preliminary results on CIFAR-100 indicate that aligning the intermediate representations of perturbed and unperturbed samples significantly enhances model performance. Figure \ref{fig:generate} shows results of the 3 methods alongside baseline training for image generation from the encoding of an image from CUB dataset. We used the latent encoding for the given image and generated 5 sample outputs, one with deterministic (i.e., without adding any noised variance) and 4 noised latents (using different noise samples for the reparameterization). Visually, there is not much difference to be seen, however the structure is more close to original in case of AdEMAMix-ICP for the deterministic outputs.

Table \ref{tab:results} show the SSIM and FID scores of VAE using the 4 methods. The results of image generation were expected to be poor due to the heavy constraints applied. However, even with such constraints, ICP still resulted in better scores as compared to the baseline. AdEMAMix-ICP performed best in terms of both SSIM and FID scores. Furthermore, even remaining 2 ICP variants still performed better than the baseline.

%% file: tables/results.tex
\begin{table}[ht]
  \centering
    \begin{tabular}{@{}lcccc@{}}
      \toprule
      \textbf{Method} & \textbf{SSIM $\uparrow$} & \textbf{FID $\downarrow$} & \makecell{\textbf{Time}\\\textbf{(mins)}$\downarrow$} \\
      \midrule
      Control      &  0.2580 & 161.830 &  \textbf{110.31}\\
      \midrule
      SGD-ICP        & 0.3365 & 159.905 & 232.52\\
      Adam-ICP       & 0.3645 & 158.08 &  221.89\\
      AdEMAMix-ICP   & \textbf{0.3893}  & \textbf{157.604} &  232.80\\
      \bottomrule
    \end{tabular}
  \caption{Quantitative evaluation on Image Generation on CUB}
  \label{tab:results}
\end{table}

%% file: sec/5_conclusion.tex
\section{Conclusion}
We evaluated our ICP-based self-distillation framework with two different settings: a vanilla classification scenario and a more restricted VAE-based generative environment. Under both cases, input refinement through ICP always helped boosting accuracy and F1 for classification and increasing SSIM while decreasing FID for image generation. These results demonstrate the power of combining input refinement with self-distillation to close the gap between fitting and generalization. While our experiments were conducted on medium-sized tasks, the framework can be extended to larger models and more intricate datasets in the future, perhaps releasing even higher performance benefits and extending its scope to real-world problems.